\documentclass[dvipdfmx]{article}

\usepackage{microtype}
\usepackage{graphicx}
\usepackage{subfigure}
\usepackage{booktabs} 
\usepackage{algorithm}

\usepackage{amsmath}
\usepackage{amssymb}
\usepackage{enumitem}
\usepackage{bm}

\DeclareMathOperator*{\argmax}{argmax}
\DeclareMathOperator*{\argmin}{argmin}

 \allowdisplaybreaks

\usepackage{hyperref}

\usepackage[accepted]{icml2018}

\icmltitlerunning{Bayesian Optimization for Multi-objective Optimization and Multi-point Search}

\begin{document}

\twocolumn[
\icmltitle{Bayesian Optimization for Multi-objective Optimization and Multi-point Search}



\icmlsetsymbol{equal}{*}

\begin{icmlauthorlist}
\icmlauthor{Takashi Wada}{KSL}
\icmlauthor{Hideitsu Hino}{ISM}
\end{icmlauthorlist}

\icmlaffiliation{KSL}{Kobe Steel, Ltd., Kobe, Japan}
\icmlaffiliation{ISM}{The Institute of Statistical Mathematics/RIKEN AIP, Tokyo, Japan}

\icmlcorrespondingauthor{Takashi Wada}{wada.takashi1@kobelco.com}

\icmlkeywords{Bayesian Optimization, Multi-objective Optimization, Multi-point Search}

\vskip 0.3in
]



\printAffiliationsAndNotice{\icmlEqualContribution} 

\begin{abstract}
Bayesian optimization is an effective method to efficiently optimize unknown objective functions with high evaluation costs. Traditional Bayesian optimization algorithms select one point per iteration for single objective function, whereas in recent years, Bayesian optimization for multi-objective optimization or multi-point search per iteration have been proposed. However, Bayesian optimization that can deal with them at the same time in non-heuristic way is not known at present. We propose a Bayesian optimization algorithm that can deal with multi-objective optimization and multi-point search at the same time. First, we define an acquisition function that considers both multi-objective and multi-point search problems. It is difficult to analytically maximize the acquisition function as the computational cost is prohibitive even when approximate calculations such as sampling approximation are performed; therefore, we propose an accurate and computationally efficient method for estimating gradient of the acquisition function, and develop an algorithm for Bayesian optimization with multi-objective and multi-point search. It is shown via numerical experiments that the performance of the proposed method is comparable or superior to those of heuristic methods.
\end{abstract}

\section{Introduction}
Performance requirements for industrial products are getting stricter, and to develop a product that satisfies the required industrial standards, it is necessary to identify optimal design conditions by repetitively evaluating performance of products through prototyping or simulation. However, expenses and time for trial productions and simulations are limited, thus it is necessary to identify optimal design conditions within a few trials.

Bayesian optimization (BO)~\citep{SHAHRIARI2016, BROCHU2010} is an efficient approach for optimizing unknown functions with high evaluation costs. BO can efficiently be used to search for a globally optimal solution $\bm{x}^{\star} = \argmin_{ \bm{x} \in \mathcal{X} } f(\bm{x})$ with respect to the unknown function $f(\bm{x})$, which represents the relation between objective variable and explanatory variable $\bm{x} \in \mathcal{X} \subset \mathbb{R}^{d_{x}}$, where  $ \mathcal{X}$ is the feasible region of $\bm{x}$. BO consists of steps of learning probability models and for determining points to be evaluated next, based on a certain evaluation criteria called {\it{acquisition function}} $J(\bm{x})$, and the global optimal solution $\bm{x}^{\star}$ is searched by repeating each step.

First, in learning probability model step, a model of the unknown function $f(\bm{x})$ is learned based on the currently available dataset $\mathcal{D}_n = \{ (\bm{x}_1, f(\bm{x}_1)), \cdots, (\bm{x}_n, f(\bm{x}_n)) \}$. A typical model for $f(\bm{x})$ is the Gaussian process (GP)~\citep{Rasmussen2006}.
Next, in the step for determining the next evaluation point, the point $\bm{x}_{n+1} = \argmax_{ \bm{x} \in \mathcal{X} } J(\bm{x})$ at which the acquisition function $J(\bm{x})$ is maximized is determined as the next point to be evaluated based on the learned probability model.
Several methods for designing the acquisition function such as probability of improvement (PI)~\citep{Torn1989}, expected improvement (EI)~\citep{Jones1998}, upper confidence bound (UCB)~\citep{Srinivas2009},
entropy search (ES)~\citep{Hernandez2015}, stepwise uncertainty reduction (SUR)~\cite{Picheny2015}, and knowledge gradient (KG)~\citep{Frazier2009} have been proposed.

A simple BO is the optimization of a single objective variable, that is modeled as an objective function $f(\bm{x})$. Also, when performing the iterative search, unknown function $f$ is evaluated in succession. However, in reality, optimizing multiple objectives in a trade-off relationship, such as strength and weight as product performance, may be required.
Also, in the evaluation phase, it is more efficient to simultaneously evaluate multiple points when performing the iterative search if it is possible to perform prototyping or simulation under multiple conditions in parallel. Owing to these requirements and circumstances, recently, BOs that can be employed in handling multi-objective optimization~\citep{Emmerich2011,Svenson2016,Ponweiser2008,Picheny2015,Hernandez2015} and multi-point searches~\citep{Ginsbourger2010, Chevalier2012, Marmin2015, 
Wu2016, Shah2015, Desautels2014, Anirban2014, Zheng2016} have been proposed. However, a BO method that can simultaneously handle both multi-objective optimization and multi-point search problems has not yet been proposed.

Hence, this paper proposes a BO method that can simultaneously handle both multi-objective optimization and multi-point searches. In the proposed method, we define an acquisition function by extending the existing multi-objective optimization and multi-point search methods. Subsequently we consider an optimization problem to find maximum of the acquisition function. The acquisition function defined here involves multivariate integration. The approximate calculation using Monte Carlo sampling is often adopted, but generally it is computationally demanding. Furthermore, na\"ive Monte Carlo sampling is not suitable for evaluating the acquisition function in BO because in a multi-point search problem the dimension of the variables to be optimized tend to significantly increase, e.g., when we consider $q$ points in a simultaneous search, it implies we have to estimate the integral with respect to $d_{x} \times q$ dimensional variables. Moreover, when the gradient of the objective function with a high-dimensional variable is estimated using sampling approximation, the approximated gradient tends to become a zero vector (vanishing gradient problem). 

Major contributions of this study are as follows:
\begin{itemize}
\item To the best of our knowledge, BO method for multi-objective and multi-point searches using non-heuristic approach is proposed for the first time in this paper.
\item We propose a computationally efficient algorithm for the proposed BO method based on Monte Carlo approximation of the gradient of the acquisition function. We empirically showed the newly designed acquisition function can effectively avoid the vanishing gradient problem.
\end{itemize}

\section{Related Work}
\label{Related Work}

\subsection{Gaussian Process}
In this work, we will use Gaussian Process (GP) as the probabilistic model for Bayesian optimization. GP is a stochastic process of a function values $f(\bm{x})$ that follows prior distribution with mean $\lambda$ and covariance between two points $f(\bm{x})$ and $f(\bm{x}')$ is defined by a positive definite kernel function $ \kappa(\bm{x}, \bm{x}^{\prime})$. We introduce the covariance matrix $K_{n}$ having elements $[K_n]_{i,j} = \kappa (\bm{x}_i, \bm{x}_j)$. 
 We adopted a well-known kernel function for GP is a Gaussian kernel of automatic relevance determination type~\citep{Rasmussen2006} for implementing the proposed method. Kernel parameters are determined by evidence maximization. 

Given the observed dataset $\mathcal{D}_n$, the posterior distribution of $f(\bm{x})$ is defined as follows:
\begin{align*}
p(f(\bm{x}) | \mathcal{D}_n) =& \mathcal{N}(\mu_n(\bm{x}), \sigma_n^2(\bm{x})),	\\
\mu_n(\bm{x}) =& \lambda + \bm{k}_n(\bm{x})^{\top} K_n^{-1} ( \bm{f} - \lambda {\bm 1} ), 	\\
\sigma_n^2(\bm{x}) =& \kappa(\bm{x},\bm{x}) - \bm{k}_n(\bm{x})^{\top} K_n^{-1} \bm{k}_n(\bm{x}),
\end{align*}
where $\bm{k}_n(\bm{x})=[\kappa(\bm{x},\bm{x}_1), \cdots, \kappa(\bm{x},\bm{x}_n)]^{\top}$, $\bm{f}=[f(\bm{x}_1), \cdots, f(\bm{x}_n)]^{\top}$, and $\bm{1}$ is a vector with all ones.
In an iterative search using BO, hyper parameters are estimated every time a new observation datum is added, and the posterior distribution $p(f(\bm{x}) | \mathcal{D}_n)$ is updated.

\subsection{Multi-objective Bayesian Optimization}
Here we describe BO dealing with multi-objective optimization that minimizes $d_f$ objective variables. Hereinafter, unknown functions expressing the relation between each objective and explanatory variable are denoted as $f^{(1)}(\bm{x}), \cdots, f^{(d_f)}(\bm{x})$. They can be collectively represented as $F(\bm{x}) = [ f^{(1)}(\bm{x}), \cdots, f^{(d_f)}(\bm{x}) ]^{\top}$. In the objective variable in the trade-off relationship, there is no single solution that minimizes each objective variable. 
For a multi-objective optimization, a solution $S_1$ is said to dominate a solution $S_2$ if all the objective values of $S_{1}$ are better than the corresponding objective values of solution $S_2$. In this case, $S_{2}$ is said to be dominated by $S_{1}$. A non-dominated solution $S$ is a solution that is not dominated by any other solution, and is also referred to as the Pareto solution. Generally, it is not a single point but a set $\mathcal{X}^\star \subset \mathcal{X}$. Also, a plane on the range of object variable formed by a set of Pareto solutions are referred to as the Pareto front.
We aim to search a finite number of solutions that closely approximate Pareto front because the Pareto solution is not a finite set. However, since the Pareto front is unknown, at the time of actual optimization, a new point is added to the surface formed by the non-dominated solution set $\mathcal{D}_n^{\star}$ defined by a set satisfying the following condition
\begin{align}
\notag
&\forall \bm{x}, (\bm{x},f(\bm{x})) \in \mathcal{D}_n^\star \subset \mathcal{D}_n,
\forall \bm{x}', (\bm{x}',f(\bm{x}')) \in \mathcal{D}_n,\\
&\exists k \in \{ 1,\cdots,d_f \}
\; \; {\rm such~that}~ f^{(k)}(\bm{x}) \le f^{(k)}(\bm{x}^{\prime})
\label{eq:pareto2}
\end{align}
based on the current observation data set $\mathcal{D}_n = \{ (\bm{x}_1, F(\bm{x}_1)), \cdots, (\bm{x}_n, F(\bm{x}_n)) \} $. The solutions are searched based on the amount of improvement when adding a new point to update the observation dataset as $\mathcal{D}_{n+1} = \{ \mathcal{D}_n, (\bm{x}_{n+1}, F(\bm{x}_{n+1})) \}$.

\begin{figure}[t]
\centering
\includegraphics[width=70mm,height=65mm]{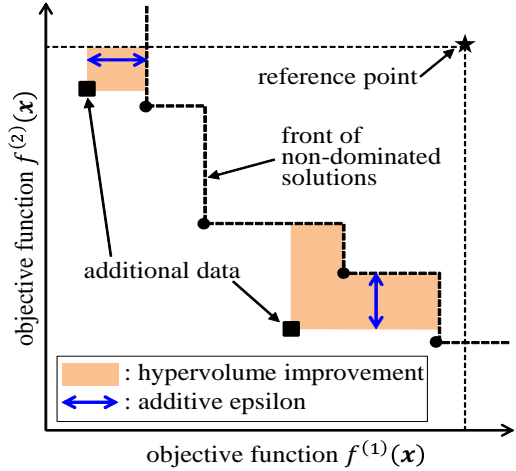}
\caption{Example of hypervolume improvement and additive epsilon for multi-objective optimization with two variables.}
\label{fig:hypervolume}
\end{figure}

Hypervolume improvement (HVI) and additive epsilon~\citep{Zitzler:2003:PAM:2221365.2221632} shown in Fig.~\ref{fig:hypervolume} are often used as measures of improvement. Suppose $HV(\mathcal{D}_n^\star)$ is the Lebesgue measure (hypervolume) of the region dominated by $\mathcal{D}_n^\star$ with a reference point (a user-defined parameter to specify the upper limit of the Pareto solution to be searched) as an upper bound. Then, HVI is defined as
\begin{align}
HVI \left( F(\bm{x}_{n+1}) \right) = HV(\mathcal{D}_{n+1}^\star) - HV(\mathcal{D}_n^\star).
\label{eq:HVI}
\end{align}

Various BO methods for multi-objective optimization have been proposed. Expected hypervolume improvement (EHI)~\cite{Emmerich2011} is an extension of EI's idea to the HVI, and the acquisition function is defined as
\begin{align}
J(\bm{x}) = \mathbb{E}_{ p \left( F(\bm{x}) \middle| \mathcal{D}_n \right) }
 \left[ HVI \left( F(\bm{x}) \right) \right],
\label{eq:EHI}
\end{align}
where $p( F(\bm{x}) |\mathcal{D}_n)$ denotes the posterior distribution of $F(\bm{x})$. To make the computation tractable, $F(\bm{x})$ is often assumed to be independent as
$p( F(\bm{x}) |\mathcal{D}_n) = p(f^{(1)}(\bm{x})|\mathcal{D}_n) \cdots p(f^{(d_f)}(\bm{x})|\mathcal{D}_n)$. Multiple output Gaussian process~\citep{pmlr-v9-alvarez10a} and BO with correlated outputs~\citep{Shah:2016:PFL:3045390.3045593} have been proposed and our proposed method can be improved by introducing the correlation, with additional computational cost. 

Expected maximum improvement (EMI) is proposed in~\citep{Svenson2016} by extending the idea of EI to the additive epsilon. \citet{Ponweiser2008} proposed the S-metric selection (SMS) method, which extends the concept of UCB to HVI. \citet{Picheny2015} proposed the SUR method, which can be used to calculate the expected value of HVI in the entire feasible region $\mathcal{X}$ and search for a point that minimizes the expected value.

\subsection{Bayesian Optimization with Multi-point Search}
Herein we present a BO method for multi-point search which searches for $q$ candidate points at each iterative search. Thereafter, $q$ candidate points $\bm{x}^{(1)}, \cdots, \bm{x}^{(q)}$ are arranged to a vector $X = [ {\bm{x}^{(1)}}^{\top}, \cdots, {\bm{x}^{(q)}}^{\top} ]^{\top}  \in \mathbb{R}^{q d_x}$.

The existing BO methods for multi-point search can be divided into two categories: The first category is a non-greedy search approach with the acquisition function $J(X)$ designed for $q$ candidate points, and the point where $X_{n+1} =  {\rm argmax}_{ X \in \mathcal{X}_q } J(X)$ is determined as the next point to be evaluated. Here, $\mathcal{X}_q$ represents a feasible region of $X$. 
Concerning the design of acquisition function $J(X)$, q-EI~\citep{Ginsbourger2010,Chevalier2012,Marmin2015} as an extension of EI, q-KG~\citep{Wu2016} as an extension of KG, and PPES~\citep{Shah2015} as an extension of ES have been proposed. 

The second category is greedy search approach. Kriging believer constant liar proposed in~\citep{Ginsbourger2010} decides the first point $\bm{x}^{(1)}$ using EI. The objective variable corresponding to the determined point $\bm{x}^{(1)}$ can be calculated using the predictive mean $\mu_n(\bm{x})$ of GP and the next point is found using EI as if a new data point ($\bm{x}^{(1)}$, $\mu_n(\bm{x})$) is obtained.
 BUCB~\citep{Desautels2014} is also a greedy search method based on UCB. It can sequentially determine candidate points while updating only the GP prediction variance $\sigma_{n}(\bm{x})$. There is also a technique of imposing a penalty to exclude neighbor points already determined in greedy search from the candidate points. For example, \citet{Anirban2014} proposed to exclude a region whose distance from the already determined points is equal to or less than a certain value.  \citet{Zheng2016} proposed to impose a penalty so as to increase the mutual information between already decided points and the next candidate.

\subsection{Heuristic Multi-objective multi-point Bayesian Optimization}
\label{HeuristicMethod}
In considering BO that can handle multi-objective optimization and multi-point searching at the same time, we will present a non-greedy method in Section~\ref{Proposed Method}. There are no other existing non-heuristic method for multi-objective and multi-point search method. Before developing a new method, we briefly discuss two na\"ive extensions of the conventional methods for multi-objective BO to multi-point search based on greedy approach, and a multi-point search method, which individually decides candidate points in succession.

The first method is based on multi-point search method proposed by~\citet{Anirban2014}, which is denoted as DC (distance constraints). In the sequential determination of the candidate points, the explanatory variables are normalized to the range of $[0, 1]$, and the Euclidean distance from the point already determined is $0.1 \sqrt{d_x}$ or less is excluded from the candidate region $\mathcal{X}$. The algorithmic description of the multi-objective and multi-point search DC algorithm is presented in the supplementary material.

The second method is based on the method proposed by~\citet{Ginsbourger2010}, which is denoted as KB (knowledge believer). For sequential determination of the candidate points, we can substitute the value of the objective variable for the already decided point with the predicted mean $\mu_n(\bm{x})$ of GP, and temporarily add it to the observation data $\mathcal{D}_n$ for use in searching for the next candidate point. The algorithmic description of the multi-objective and milt-point search KB algorithm is presented in the supplementary material.

In both these methods, the optimization of the acquisition function is done by using a conventional BO method for multi-objective functions.

\section{Proposed Method}
\label{Proposed Method}
In this section, we present the proposed non-greedy multi-point search method based on the concept of EHI in multi-objective optimization.

\subsection{Design of Acquisition Function}

As a simple extension of the acquisition function of EHI in Eq.~\eqref{eq:EHI} to a non-greedy multi-point search, we define the following acquisition function
\begin{align}
q \mathchar`- EHI(X) = 
 \mathbb{E}_{ p ( F_q(X) | \mathcal{D}_n ) }
 \left[ HVI \left( F_q(X) \right) \right],
\label{eq:qEHI}
\end{align}
where 
$F_q(X) = [ F(\bm{x}^{(1)})^{\top}, \cdots, F(\bm{x}^{(q)})^{\top} ]^{\top}  \in \mathbb{R}^{qd_f}$, 
$HVI ( F_q(X) )$ is the improvement of the hypervolume when $q$ candidates $F(\bm{x}^{(1)}), \cdots, F(\bm{x}^{(q)})$ are added to $\mathcal{D}_n^{\star}$. 
$p ( F_q(X) | \mathcal{D}_n )$ is the posterior of $F_{q}(X)$ given the observation dataset $\mathcal{D}_n$. Each of the objective functions $f^{(k)}(\bm{x})$ are assumed to follow independent Gaussian processes as $ p \left(F_q(X)  | \mathcal{D}_n \right) = p \left( f^{(1)}(\bm{x}^{(1)}), \cdots, f^{(1)}(\bm{x}^{(q)}) | \mathcal{D}_n \right) \cdots $
$ p \left( f^{(d_f)}(\bm{x}^{(1)}), \cdots, f^{(d_f)}(\bm{x}^{(q)}) | \mathcal{D}_n \right)$. 

\begin{figure}[t]
\centering
\includegraphics[width=70mm,height=65mm]{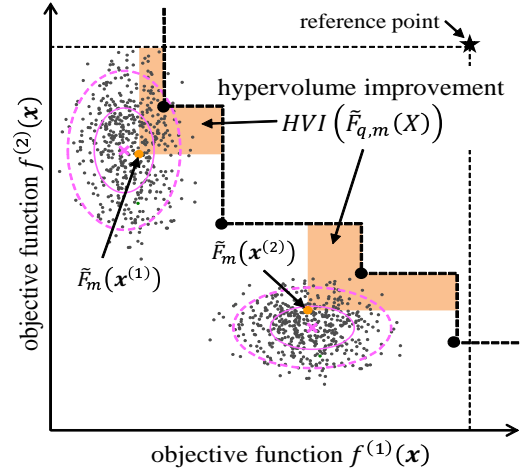}
\caption{Conceptual diagram of approximation of the acquisition function $q \mathchar`- EHI$ using Monte Carlo sampling. $HVI(\tilde{F}_{q,m}(X))$ indicates the hypervolume improvement by the $m$-th sample point $\tilde{F}_{m}(\bm{x}^{(1)})$ and $\tilde{F}_{m}(\bm{x}^{(2)})$.}
\label{fig:qEHI_MC}
\end{figure}

Since the acquisition function~\eqref{eq:qEHI} requires an integral with respect to multivariate distribution for calculating expectation, it is difficult to calculate it analytically. Indeed, it is pointed out the by \citet{Hernandez2015} that computation of $EHI$ for multi-objective problem is feasible at most for two or three objectives, and when we consider multi-point search, the difficulty of the computation would exponentially grow. Therefore, approximation using Monte Carlo sampling is performed:
\begin{align}
q \mathchar`- EHI_{MC}(X) = 
 \frac{1}{M} \sum_{m=1}^{M}  HVI \left( \tilde{F}_{q,m}(X) \right),
\label{eq:qEHI_MC}
\end{align}
where $M$ is the number of Monte Carlo samplings, $ \tilde{F}_{q,m}(X)$ is the $m$-th sample point that follows the distribution $p (F_q(X)  | \mathcal{D}_n )$. 
As an example, let us consider a case $d_f=2, q=2$ as shown in Fig.~\ref{fig:qEHI_MC}. In this case, the elements of $\tilde{F}_{q,m}(X)$ are
$ \tilde{F}_{2,m}(X) 
 = [ \tilde{F}_{m}(\bm{x}^{(1)})^{\top}, \tilde{F}_{m}(\bm{x}^{(2)})^{\top} ]^{\top}
 = [ \tilde{f}_{m}^{(1)}(\bm{x}^{(1)}), \tilde{f}_{m}^{(2)}(\bm{x}^{(1)}), \tilde{f}_{m}^{(1)}(\bm{x}^{(2)}), \tilde{f}_{m}^{(2)}(\bm{x}^{(2)}) ]^{\top}$. 

To find the maximizer of the acquisition function, it is possible to use metaheuristics such as a genetic algorithm. 
However, because an approximate evaluation of the acquisition function itself is computationally intensive, and the dimension of variable to be optimized can be very high when  multiple point search is considered, metaheuristics which requires evaluating the objective function many times is prohibitive. Therefore, we derive a method for approximating the gradient of the acquisition function for efficient optimization based on the gradient method.

\subsection{Gradient of the Acquisition Function}
\label{subse:dqEHI}

The gradient of the acquisition function~\eqref{eq:qEHI_MC} is given by
\begin{align}
\nabla q \mathchar`- EHI_{MC}(X) = 
 \frac{1}{M}
 \sum_{m=1}^{M}  \frac{ \partial HVI \left( \tilde{F}_{q,m}(X) \right) }
 { \partial X },
\label{eq:dqEHI_MC}
\end{align}
where
\begin{align}
\frac{ \partial HVI \left( \tilde{F}_{q,m}(X) \right) } { \partial X } =
 \begin{bmatrix}
 \frac{ \partial HVI \left(\tilde{F}_{q,m}(X) \right) } { \partial \bm{x}^{(1)} }	\\	
 \vdots 	\\
 \frac{ \partial HVI \left( \tilde{F}_{q,m}(X) \right) } { \partial \bm{x}^{(q)} }
 \end{bmatrix}
\label{eq:dqEHI_MC2}
\end{align}
\begin{align*}
\frac{ \partial HVI \left( \tilde{F}_{q,m}(X) \right) } { \partial \bm{x}^{(i)} }
 &=  \frac{ \partial \tilde{F}_{q,m}(\bm{x}^{(i)}) } { \partial \bm{x}^{(i)} }
     \frac{ \partial HVI \left( \tilde{F}_{q,m}(X) \right) } { \partial \tilde{F}_{q,m}(\bm{x}^{(i)}) }		\notag \\
 &= \sum_{k=1}^{d_f} \frac{ \partial HVI \left( \tilde{F}_{q,m}(X) \right) } { \partial \tilde{f}^{(k)}_{m}(\bm{x}^{(i)}) }
 \frac{ \partial \tilde{f}^{(k)}_{m}(\bm{x}^{(i)}) } { \partial \bm{x}^{(i)} }.
\end{align*}
In order to calculate this gradient, we need to calculate the partial derivatives ${ \partial HVI (  \tilde{F}_{q,m}(X) ) }/{ \partial \tilde{f}^{(k)}_{m}(\bm{x}^{(i)}) }$ and 
${ \partial \tilde{f}^{(k)}_{m}(\bm{x}^{(i)}) }/{ \partial \bm{x}^{(i)} }$. 

Concerning the derivative ${ \partial HVI (  \tilde{F}_{q,m}(X) ) }/{ \partial \tilde{f}^{(k)}_{m}(\bm{x}^{(i)}) }$, 
in actual problems requiring multi-objective optimization, $d_{f}$ tends to be small, thus numerical derivation approach is computationally tractable in such scenario. 
Let $\delta_{k}$ be a vector with a small positive number at the $k$-th element and zero otherwise. We can approximate the infinitesimal change in $HVI(\tilde{F}_{q,m}(X))$ by $HVI ( \tilde{F}_{q,m}(X) + \delta_k ), k=1,\cdots,d_f$ caused by an infinitesimal change in $\tilde{F}_{q,m}(X)$, and perform numerical differentiation using $d_{f}+1$-times evaluation of the $HVI$.

On the other hand, ${ \partial \tilde{f}^{(k)}_{m}(\bm{x}^{(i)}) }/{ \partial \bm{x}^{(i)} }$ cannot be estimated by numerical derivation in the same manner as ${ \partial HVI (  \tilde{F}_{q,m}(X) ) }/{ \partial \tilde{f}^{(k)}_{m}(\bm{x}^{(i)}) }$, because for each $q \times d_{x}$ element of $X$, we need to sample perturbed points according to the probability $p ( F_q(X) | \mathcal{D}_n )$. In general, $d_{x} > d_{f}$ and the number of points to be searched $q$ is set to a maximum allowable number by the prototyping or simulation system. The computational cost of estimating the probability distribution for a new point given small perturbation to an element of $X$ is of order $\mathcal{O}(n^{2})$ and it grows in the process of BO.

One of the major contributions of this work is in deriving a computationally efficient method for estimating ${ \partial \tilde{f}^{(k)}_{m}(\bm{x}^{(i)}) }/{ \partial \bm{x}^{(i)} }$, by generating sample gradients of $\tilde{f}^{(k)}_m(\bm{x})$ that follows the distribution $p ( f^{(k)}(\bm{x}) | \mathcal{D}_n )$. Because $\tilde{f}^{(k)}_m(\bm{x})$ cannot be described as an explicit function of $\bm{x}$, we introduce an approximation method. For a shift-invariant kernel $\kappa$, from the Bochner's theorem~\citep{Bochner1957}, there exists a Fourier dual $s(\bm{w})$ as the spectral density of $\kappa$ and for the normalized probability density function $p(\bm{w}) = s(\bm{w}) / \alpha$ and a realization $b$ of uniform random variable $B \sim U[0, 2\pi]$, we have
\begin{align*}
\kappa (\bm{x}, \bm{x}^{\prime}) 
 &= \alpha \mathbb{E}_{p(\bm{w})} 	\left[ \exp \left( -j \bm{w}^{\top} ( \bm{x} - \bm{x}' ) \right) \right]      \notag \\
 &= 2 \alpha \mathbb{E}_{p(\bm{w},b)} \left[ \cos (\bm{w}^{\top} \bm{x}+b) \cos (\bm{w}^{\top} \bm{x}^{\prime}+b) \right].
\end{align*}
Let $W \in \mathbb{R}^{r \times d_x}, \bm{b} \in \mathbb{R}^{r}$ be the $r$ realizations sampled from $p(\bm{w},b)$, and consider a basis function ${\bm \phi} (\bm{x}) = \sqrt{2 \alpha / r} \cos ( W \bm{x} + \bm{b} ) \in \mathbb{R}^{r},
$ where $\cos$ acts element-wise. Then, the value of the kernel function is approximated as $\kappa (\bm{x}, \bm{x}^{\prime}) \simeq {\bm \phi} (\bm{x})^{\top} {\bm \phi} (\bm{x}^{\prime})$. Now the sample drawn from probability distribution $p ( f^{(k)}(\bm{x}) | \mathcal{D}_n )$ is approximated by a linear model
\begin{align}
g^{(k)}(\bm{x}) =& {\bm \phi} (\bm{x})^{\top} {\bm \theta_{\phi}} + \lambda^{(k)},
\label{eq:Bochnersappro}	\\
{\bm \theta_{\phi}} =& \left( \Phi^{\top} \Phi \right)^{-1} \Phi^{\top}(\bm{f}^{(k)} - \lambda^{(k)} {\bm 1}),
\label{eq:prob_theta}
\end{align}
where $\Phi = [ {\bm \phi}(\bm{x}_1), \cdots, {\bm \phi}(\bm{x}_n) ]^{\top}$, and $\lambda^{(k)}$ is the expectation of the prior distribution of the $k$-th objective variable. It is guaranteed that probability distribution of $g^{(k)}(\bm{x})$ converges to $p ( f^{(k)}(\bm{x}) | \mathcal{D}_n )$ as $r \to \infty$ ~\citep{Neal1995}.

Equation~\eqref{eq:Bochnersappro} is a function of $\bm{x}$. Let $[W]_{:,i}$ be the $i$-th column vector of $W$. The gradient of $g^{(k)}$ is given by
\begin{align}
&\frac{\partial g^{(k)}(\bm{x})}{\partial \bm{x}[i]} 
 \!=\! { \frac{\partial {\bm \phi}(\bm{x})}{\partial \bm{x}[i]} }^{\top} {\bm \theta_{\phi}}	\notag \\
 &\!=\! \left( -\sqrt{\frac{2\alpha}{r}} {\rm diag}([W]_{:,i}) \sin (W \bm{x} + \bm{b}) \right)^{\top} \! {\bm \theta_{\phi}}.
\label{eq:dBochnersappro}
\end{align}
Now we can efficiently estimate the gradient of the acquisition function~\eqref{eq:dqEHI_MC}. 
\begin{figure}[t]
\centering
\includegraphics[width=70mm,height=65mm]{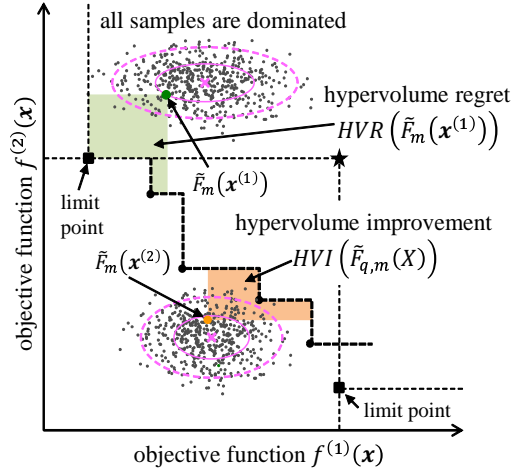}
\caption{The acquisition function with regret $HVR$ to deal with the vanishing gradient problem. All of the sample points correspond to $\bm{x}^{(1)}$ are dominated and the regret $HVR(\tilde{F}_{m}(\bm{x}^{(1)}))$ shows the hypervolume dominated by the set of non-dominated solutions and limit points (a user-defined parameter for specifying the lower limit of the value of the objective variable) with the reference point $\tilde{F}_{m}(\bm{x}^{(1)})$.}
\label{fig:qEHIR}
\end{figure}

\subsection{Dealing with the Vanishing Gradient Problem}

Even when the candidate point $X$ has a small probability of improvement, the value of the acquisition function of Eq.~\eqref{eq:qEHI} can have a positive value $q \mathchar`- EHI > 0$, and an improvement direction of $q \mathchar`- EHI$ exists.
However, when calculating the acquisition function of Eq.~\eqref{eq:qEHI_MC} approximately by Monte Carlo sampling, all sample points tend to be dominated and $q \mathchar`- EHI_{MC}(X) = 0$. As a result, $\nabla q \mathchar`- EHI_{MC}(X) = 0$, and the gradient-based optimization could get disrupted. Here, this problem is referred to as the {\it{vanishing gradient problem}}. To solve this problem, we introduce the idea of {\it{regret}} as shown in Fig.~\ref{fig:qEHIR}. 

First we define the regret $HVR_{MC}^{(q^{\prime})}$ at a candidate point $\bm{x}^{(q^{\prime})}$ as
\begin{flalign}
HVR_{MC}^{(q^{\prime})} =
\begin{cases}
\frac{1}{M} \sum_{m=1}^{M}  HVR \left( \tilde{F}_{m}(\bm{x}^{(q^{\prime})}) \right)  \\
  (\tilde{F}_{m}(\bm{x}^{(q^{\prime})}), \forall m \; \;\mbox{are dominated} )  \\
 0 \hspace{4mm} ({\rm otherwise}).
\end{cases}
\label{eq:HVRq}
\end{flalign}
In this definition, $HVR ( \tilde{F}_{m}(\bm{x}^{(q^{\prime})}) )$ is the hypervolume dominated by the non-dominated solution set $\mathcal{D}_n^{\star}$ with a reference point $\tilde{F}_{m}(\bm{x}^{(q^{\prime})})$ and the predefined limit points. The regret is an index indicating the extent the candidate point is dominated, and it becomes larger as the probability of improvement is smaller. Now we introduce a novel acquisition function by subtracting regret~\eqref{eq:HVRq} from~\eqref{eq:qEHI_MC} as
\begin{align}
q \mathchar`- EHIR_{MC}(X) 
=& \frac{1}{M} \sum_{m=1}^{M}  HVI \left( \tilde{F}_{q,m}(X) \right)	\notag \\
 &- \sum_{q^{\prime}=1}^{q}  HVR_{MC}^{(q^{\prime})}.
\label{eq:qEHIR_MC}
\end{align}
As shown in Fig.~\ref{fig:qEHIR}, when there exists $q^{\prime}$ such that $HVR_{MC}^{(q^{\prime})} \neq 0$, i.e., when there are candidate points with a very low probability of improvement, this value decreases due to the penalty of regret with respect to the candidate point $\bm{x}^{(q^{\prime})}$. Accordingly, when maximizing the acquisition function, the effect of reducing the amount of regret is imposed, and it is expected that the candidate point $\bm{x}^{(q^{\prime})}$ will move towards improvement. It should be noted that when $HVR_{MC}^{(q^{\prime})} = 0$ at all candidate points, this novel acquisition function reduces to Eq.~\eqref{eq:qEHI_MC}. 

The gradient of the acquisition function~\eqref{eq:qEHIR_MC} is
\begin{align}
\nabla q \mathchar`- EHIR_{MC}(X) 
 =& \frac{1}{M}
 \sum_{m=1}^{M}  \frac{ \partial HVI \left( \tilde{F}_{q,m}(X) \right) }
 { \partial X }	 \notag \\
 &-  \sum_{q^{\prime}=1}^{q}  \frac{ \partial HVR_{MC}^{(q^{\prime})} }
 { \partial X }.
\label{eq:dqEHIR_MC}
\end{align}
The first term is nothing but Eq.~\eqref{eq:dqEHI_MC}. The second term is, as in Subsection~\ref{subse:dqEHI}, divided into the partial derivative of $HVR$ with respect to the objective variable and that with respect to the explanatory variable. The former is calculated by using numerical differentiation while the latter is estimated using Eq.~\eqref{eq:dBochnersappro}. We summarized the proposed method, which is referred to as multi-objective multi-point Bayesian optimization (MMBO), in Algorithm~\ref{alg:MMBO}.
\begin{algorithm}[t!]
\caption{Multi-objective multi-point Bayesian optimization (MMBO)}
\label{alg:MMBO}
\begin{algorithmic}[1]
\REQUIRE initial data $\mathcal{D}_0$ 
\WHILE{stopping criteria is not satisfied}
		\STATE	fit $d_f$ GP models based on $\mathcal{D}_{n-1}$.
		\STATE	sample $W, \bm{b} \sim p(\bm{w},b)$ and generate $M$ random functions $g^{(k)}_m$ and calculate gradient functions ${\partial g^{(k)}_m} / {\partial \bm{x}[i]}, ( m=1,\cdots,M )$ for each GP model $k=1,\dots,d_{f}$.
	\STATE	select $X = {\rm argmax}_{ X \in \mathcal{X}_q } 
			q \mathchar`- EHIR_{MC}(X)$
	by gradient method using $\nabla q \mathchar`- EHIR_{MC}(X)$.
	\STATE evaluate objective functions at $\bm{x}^{(1)}, \cdots, \bm{x}^{(q)}$ to obtain 
	 $F(\bm{x}^{(1)}), \cdots, F(\bm{x}^{(q)})$.
	\STATE augment data $\mathcal{D}_n = \mathcal{D}_{n-1} \cup \{ (\bm{x}^{(1)}, F(\bm{x}^{(1)})), \cdots, $  
	 \hspace{4mm} $(\bm{x}^{(q)}, F(\bm{x}^{(q)})) \}$.   
\ENDWHILE
\end{algorithmic}
\end{algorithm}

\section{Experimental Results}
\label{Example}
To show the effectiveness of MMBO, we will first show its calculation cost and approximation accuracy in the gradient calculation of the acquisition function as discussed in Section~\ref{subse:dqEHI}. Subsequently, we compare MMBO and the heuristic method discussed in Subsection~\ref{HeuristicMethod} on the test functions introduced in~\citep{Huband2006}.

\subsection{Evaluation of the Gradient Approximation}
\label{subse:time accur}

\begin{table*}[h!]
\centering
\caption{Characteristics of test functions and parameters in the experiment.}
\begin{tabular}{lcccccl} 									\toprule
Name	& $d_x$	& $d_f$	& Modality		& Geometry 		& Reference point		& Limit points	\\ \midrule
ZDT1	& 6	 	& 2		& uni-modal	& convex			& $[1,1]$		& $[-1,1], [1,-1]$	\\
ZDT3	& 3		& 2		& multi-modal	& disconnected	& $[1,1]$		& $[-1,1], [1,-2]$	\\
DTLZ2	& 6		& 3		& uni-modal 	& concave			& $[1,1,1]$		& $[-1,-1,1], [-1,1,-1], [1,-1,-1]$	\\ \bottomrule
\end{tabular}
\label{table:testfunc}
\end{table*}

\begin{figure}[t!]
\centering
\includegraphics[width=80mm,height=56mm]{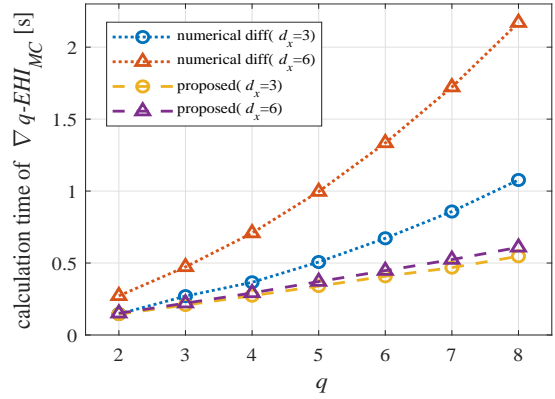}
\caption{Computational costs of MMBO and that of numerical derivative for calculating the gradient of the acquisition function for ZDT1 with $d_x=3,6$ and $q=2,\dots,8$.}
\label{fig:calc_qEHIR}
\end{figure}
\begin{figure}[t!]
\centering
\includegraphics[width=80mm,height=56mm]{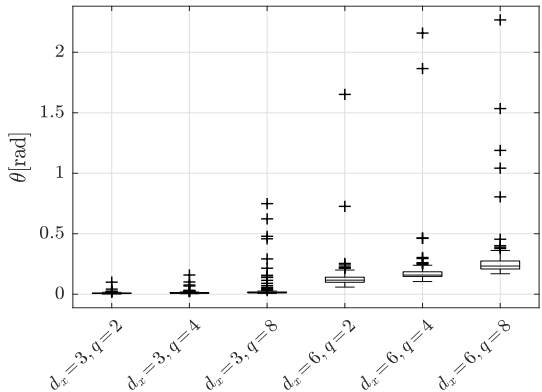}
\caption{Accuracy of the estimate of $\nabla q \mathchar`- EHI_{MC}$ by MMBO for the test function ZDT1 when $d_x=3,6$ and $q=2,4,8$.}
\label{fig:accur_qEHIR}
\end{figure}

\begin{figure*}[h!]
\begin{tabular}{ccc}
\begin{minipage}[t]{0.32\hsize}
\centering
\includegraphics[width=50mm,height=42mm]{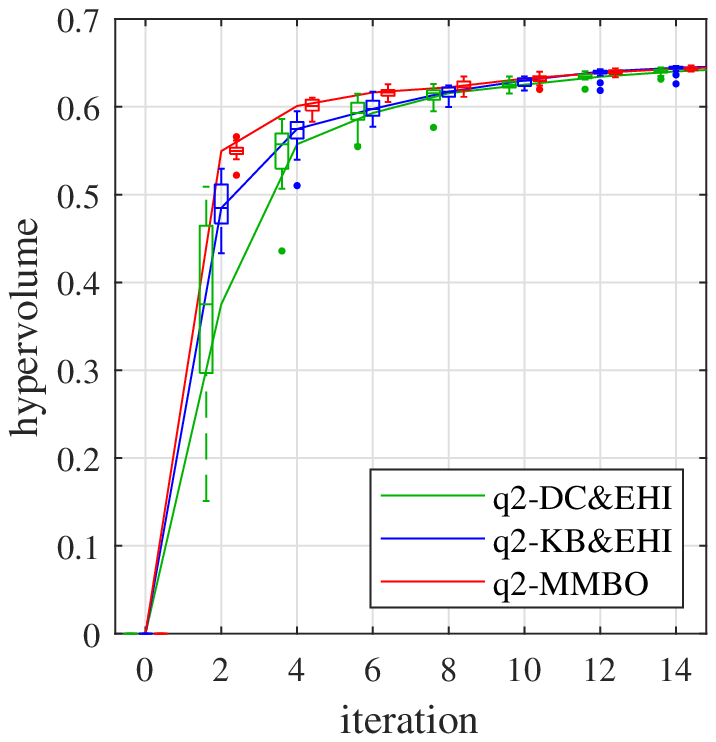}
\end{minipage} &
\begin{minipage}[t]{0.32\hsize}
\centering
\includegraphics[width=50mm,height=42mm]{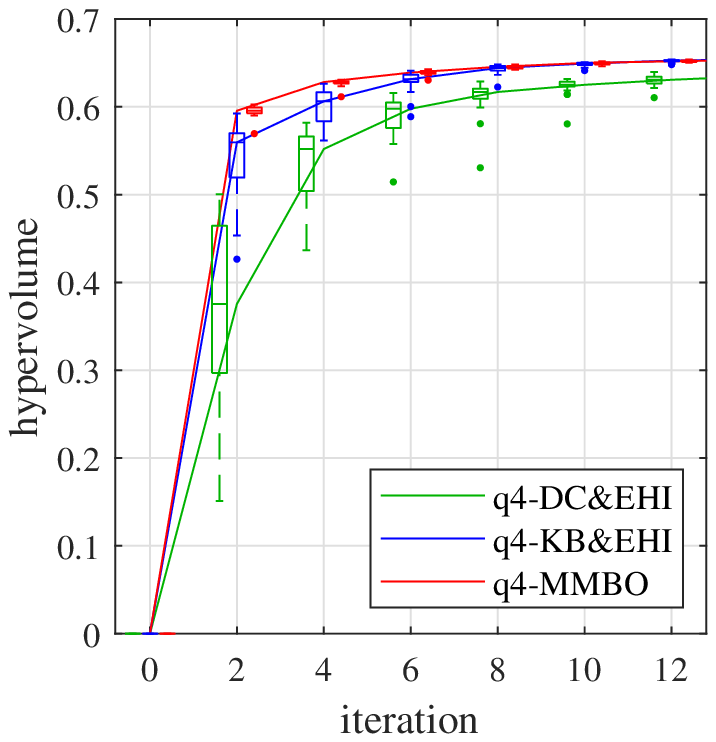}
\end{minipage} &
\begin{minipage}[t]{0.32\hsize}
\centering
\includegraphics[width=50mm,height=42mm]{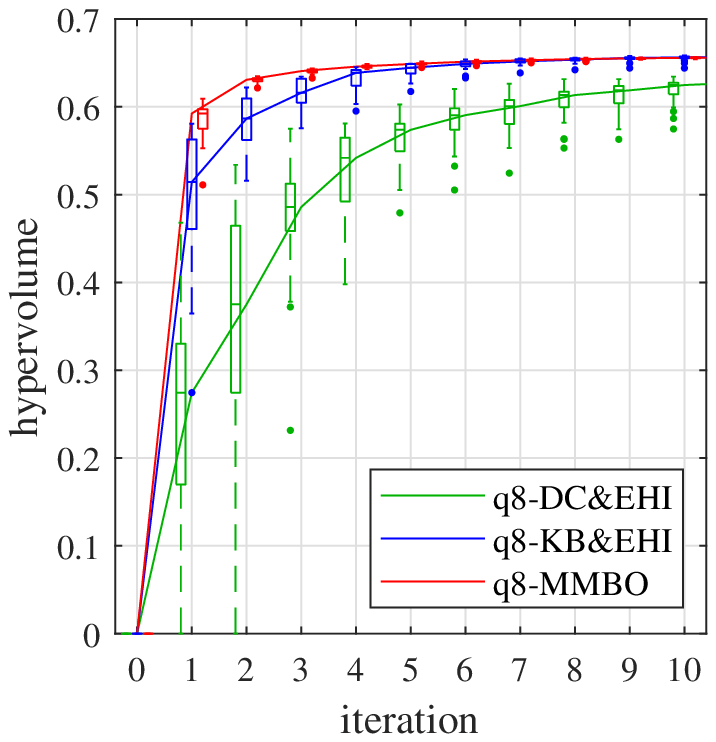}
\end{minipage}\\
(a) ZDT1, $q=2$. & (b) ZDT1, $q=4$. & (c) ZDT1, $q=8$.
\end{tabular}
\begin{tabular}{ccc}
\begin{minipage}[t]{0.32\hsize}
\centering
\includegraphics[width=50mm,height=42mm]{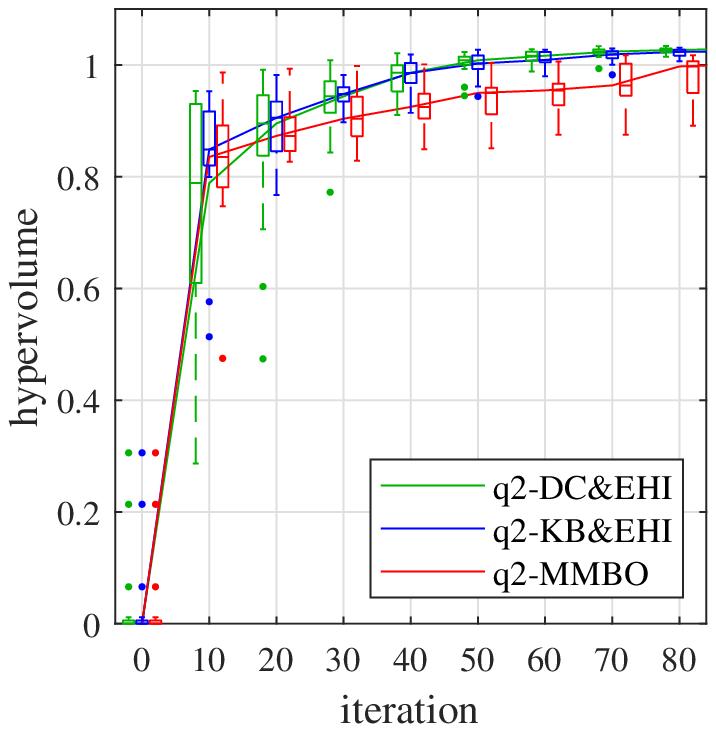}
\end{minipage} &
\begin{minipage}[t]{0.32\hsize}
\centering
\includegraphics[width=50mm,height=42mm]{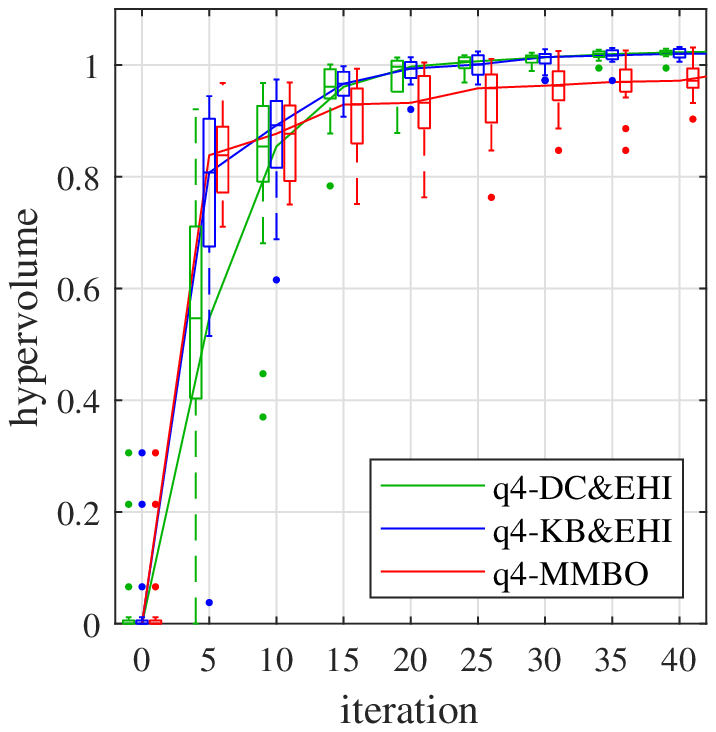}
\end{minipage} &
\begin{minipage}[t]{0.32\hsize}
\centering
\includegraphics[width=50mm,height=42mm]{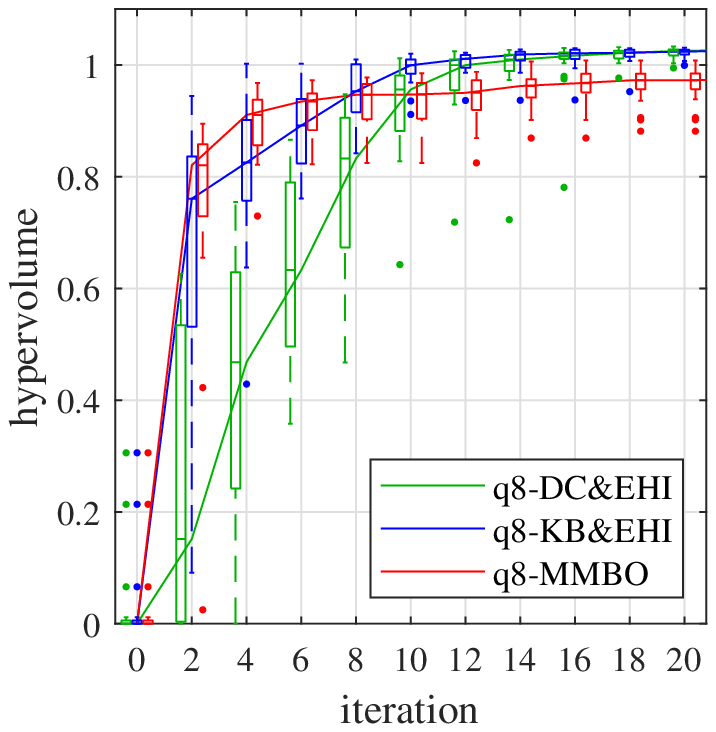}
\end{minipage}\\
(d) ZDT3, $q=2$. & (e) ZDT3, $q=4$. & (f) ZDT3, $q=8$.
\end{tabular}
\begin{tabular}{ccc}
\begin{minipage}[t]{0.32\hsize}
\centering
\includegraphics[width=50mm,height=42mm]{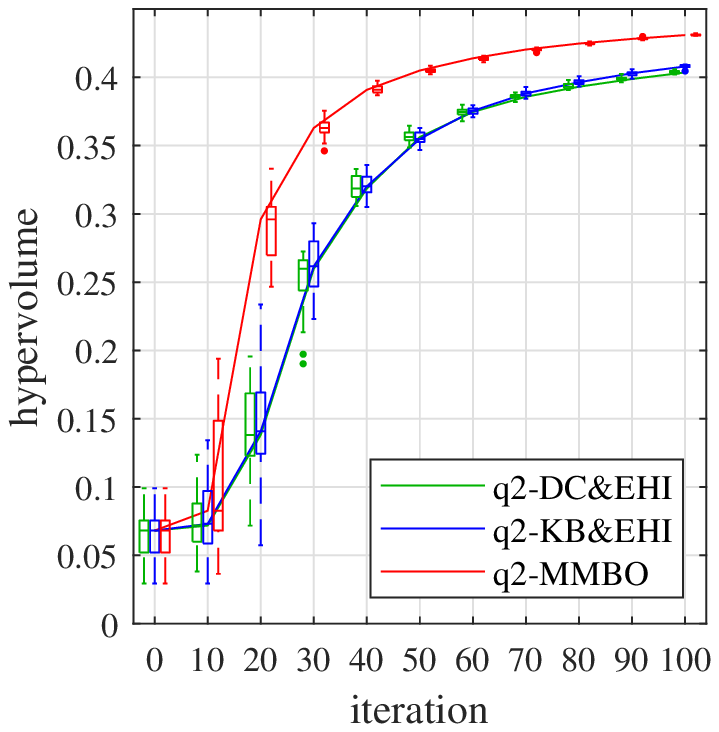}
\end{minipage} &
\begin{minipage}[t]{0.32\hsize}
\centering
\includegraphics[width=50mm,height=42mm]{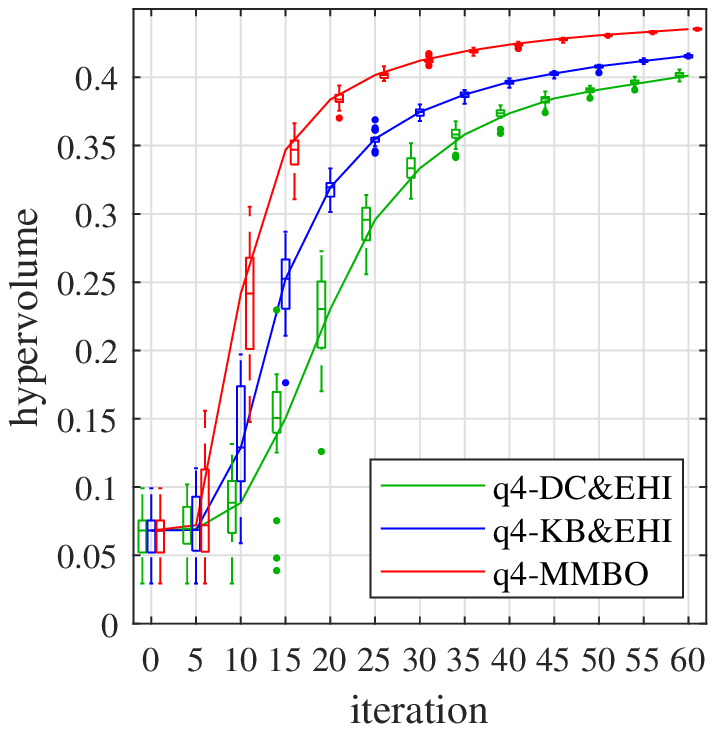}
\end{minipage} &
\begin{minipage}[t]{0.32\hsize}
\centering
\includegraphics[width=50mm,height=42mm]{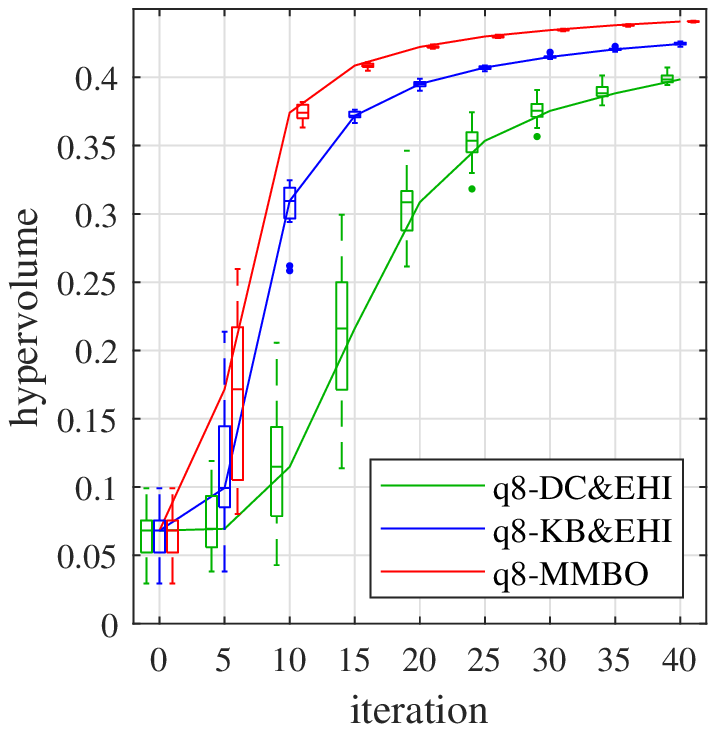}
\end{minipage}\\
(g) DTLZ2, $q=2$. & (h) DTLZ2, $q=4$. & (i) DTLZ2, $q=8$.
\end{tabular}
\caption{Comparison of MMBO and heuristic methods on test functions ZDT1, ZDT3, and DTLZ2.}
\label{fig:DTLZ2}
\end{figure*}

We evaluate the computational cost and approximation accuracy of gradient of acquisition function by MMBO on the test function ZDT1~\citep{Huband2006}. ZDT1 is a test function that can set the number of arbitrary explanatory variables with two objective variables $(d_f=2)$, and $d_{x} = 3, 6$. We also varied the number of multi-points to be searched as $q$ from $2$ to $8$.

The number of samples for approximating the acquisition function~(\ref{eq:qEHIR_MC}) is set to $M=200$, and the number of bases $\bm{\phi}(\bm{x})$ for approximating the posterior of GP is set to $r=300$. The number of initial observation was set to $n=30$ and we considered the approximate computation of the gradient of the acquisition function at a random point.

Figure~\ref{fig:calc_qEHIR} shows a comparison result of the computational costs between the proposed method and those by using numerical differentiation of the acquisition function. The computational cost increases as the dimension $d_{x}$ and the number of search points $q$ increase. The increase in computational cost for a numerical differentiation-based method with the increase in $d_x$ and $q$ is rapid while that of MMBO is relatively low. It is noteworthy that an increase in the dimension of the explanatory variable has a minor impact on MMBO. This is because the computational cost of the proposed gradient approximation method is mainly associated with the evaluation of the numerical derivative of the acquisition function by the objective function, which is independent of the dimension of the explanatory variable.

Next we evaluate the accuracy of the gradient approximation. Since it is difficult to calculate the ground truth of the gradient, we performed numerical derivative with $M=1,000$ and regard the result as the (nearly) ground truth gradient. Then, we calculate the angle $\theta$ between the ground truth and the estimated gradient using the proposed method. Figure~\ref{fig:accur_qEHIR} shows the accuracy of the gradient estimation in $\theta$ [rad] when the dimension of the explanatory variable and the number of points to be searched are varied. For each combination of the dimension of the explanatory variable and number of points to be searched, we randomly sampled $100$ points and gradients were evaluated at these points (Fig.~\ref{fig:accur_qEHIR}). From this figure, it can be seen that lower the dimension and the smaller the search points, higher the accuracy of the gradient estimation method. It can also be seen that even when $d_x=6, q=8$, the average of the angle between the true and estimated gradients is $\bar{\theta}=0.30$[rad], indicating high accuracy of the MMBO method. 

From these results, we can conclude that the proposed method can efficiently approximate the gradient of the acquisition function and is computationally efficient.


\subsection{Evaluation of the Search Efficiency}
We evaluated the search efficiency of the multi-point search method with multi-objective optimization methods using three test functions. Each of these test functions is characterized by the dimension of explanatory variable $d_{x}$, the number of objective functions $d_{f}$, the shape of the objective functions (unimodal/multi-modal), and the shape of Pareto front (concave/convex/disconnected), as shown in Table~\ref{table:testfunc}.

The parameters of the MMBO are the same as those discussed in Subsection~\ref{subse:time accur}. Quasi-Newton method is used for optimizing the acquisition function with the approximated gradient using the proposed method. The number of randomly selected initial data points is $n=3 d_x$. The quality of the search is evaluated by the hypervolume dominated by the observed data at each iteration of the optimization process. We compared MMBO with the two heuristic methods discussed in Subsection~\ref{HeuristicMethod}. In these two heuristic methods, the acquisition function was derived from EHI in the same manner as in our proposed method, and we used the genetic algorithm for optimizing the acquisition function.

By changing the random number for sampling the initial observation points $20$ times, we evaluate the hypervolumes at each step of iteration. Figure~\ref{fig:DTLZ2} shows the search results for each test function when the number of search points $q$ was set at $q=2,4,8$. 
For test functions ZDT1 and DTLZ2, the hypervolumes converged to a large value in the early stage of the optimization in MMBO, and shown the best search efficiency among all the methods. Conversely, for ZDT3, the value to which the final hypervolume converged was inferior to those obtained in the case of heuristic methods. This result can be attributed to the fact that the test function ZDT3 is multi-modal and is difficult for gradient-based methods to find the global optimal solution. We discovered that at the early stage of the optimization for ZDT3, MMBO was comparable to other heuristic methods, and for $q=8$ in which the number of search points is large, MMBO outperformed the others. 

\section{Conclusion}
\label{Conclusion}
We proposed a novel non-heuristic Bayesian optimization method that can simultaneously handle multi-objective optimization and multi-point searches. We defined an acquisition function that can simultaneously handle multi-objective optimization and multi-point search problems, and proposed an efficient and  accurate method for calculating the gradient of the acquisition function.
 The effectiveness of the proposed method was validated using different test functions and it was confirmed that it can be searched more efficiently than the heuristic methods for the unimodal objective function.
 Conversely, we found that the proposed method is subject to local solution problem. Our future work will focus on better ways of finding a good initial value in the gradient method and improving the acquisition function itself so as not to fall into the local solution. Furthermore, we proposed an approximation method for calculating the gradient of the acquisition function. Our future work will include combining the proposed approximation method with the recently developed quasi-Monte Carlo sequence-based methods~\citep{Leo2014,Buchholz2018} or quadrature Fourier feature-based method~\citep{NIPS2018_8115} for stochastic gradients to improve its convergence speed.

\bibliography{paper}
\bibliographystyle{icml2018}

\end{document}


\twocolumn[
\icmltitle{Supplementary Material for "Bayesian Optimization for Multi-objective Optimization and Multi-point Search"}




\begin{icmlauthorlist}
\icmlauthor{Takashi Wada}{KSL}
\icmlauthor{Hideitsu Hino}{ISM}
\end{icmlauthorlist}

\icmlaffiliation{KSL}{Kobe Steel, Ltd., Kobe, Japan}
\icmlaffiliation{ISM}{The Institute of Statistical Mathematics/RIKEN AIP, Tokyo, Japan}

\icmlcorrespondingauthor{Takashi Wada}{wada.takashi1@kobelco.com}

\icmlkeywords{Bayesian Optimization, Multi-objective Optimization, Multi-point Search}

\vskip 0.3in
]




\section{Algorithmic Description of Heuristic Methods}
The algorithmic details for two heuristic methods for multi-objective and multi-point search Bayesian optimization, which are n\"ive extensions of existing methods, are given as Algorithm~2 and Algorithm~3.
\begin{algorithm}[h]         
\caption{Heuristic Method : Distance Constraints}
\label{alg:Dis}
\begin{algorithmic}[1]
\REQUIRE initial data $\mathcal{D}_0$ 
\WHILE{stopping criteria is not satisfied}
	\STATE fit $d_f$ GP models based on $\mathcal{D}_{n-1}$.
	\FOR{$q'=1, \cdots, q$}
		\STATE	select $\bm{x}^{(q')} = {\rm argmax}_{ \bm{x} \in \mathcal{X} } J(\bm{x}) \quad$
		 	 s.t. $\quad{\rm min} \{ | \bm{x} - \bm{x}^{(i)} | : i=1,\cdots,q'-1 \} \geq 0.1 \sqrt{d_x} $.
	\ENDFOR
	\STATE evaluate objective functions at $\bm{x}^{(1)}, \cdots, \bm{x}^{(q)}$ to obtain 
	  $F(\bm{x}^{(1)}), \cdots, F(\bm{x}^{(q)})$.
	\STATE augment data $\mathcal{D}_n = \mathcal{D}_{n-1} \cup \{ (\bm{x}^{(1)}, F(\bm{x}^{(1)})), \cdots, $  
	   $(\bm{x}^{(q)}, F(\bm{x}^{(q)})) \}$. 
\ENDWHILE
\end{algorithmic}
\end{algorithm}
\newpage
\begin{algorithm}[t!]
\caption{Heuristic Method : Kriging Believer}
\label{alg:KB}
\begin{algorithmic}[1]
\REQUIRE initial data $\mathcal{D}_0$ 
\WHILE{stopping criteria is not satisfied}
	\FOR{$q'=1, \cdots, q$}
		\IF{$q'=1$} 
			\STATE $\hat{\mathcal{D}}_{q'} = \mathcal{D}_{n-1}$
		\ELSE
			\STATE $\hat{\mathcal{D}}_{q'} = \hat{\mathcal{D}}_{q'-1} \cup \{ (\bm{x}^{(q'-1)}, \hat{F}(\bm{x}^{(q'-1)}) ) \} $
		\ENDIF
		\STATE	fit $d_f$ GP models based on $\hat{\mathcal{D}}_{q'}$.
		\STATE	select $\bm{x}^{(q')} = {\rm argmax}_{ \bm{x} \in \mathcal{X} } J(\bm{x})$.
		\STATE 	predict objective functions at $\bm{x}^{(q')}$ to obtain 
		 $ \hat{F}(\bm{x}^{(q')}) = [ \mu_n^{(1)}(\bm{x}^{(q')}), \cdots, \mu_n^{(d_f)}(\bm{x}^{(q')}) ]^{\top} $.
	\ENDFOR
	\STATE evaluate objective functions at $\bm{x}^{(1)}, \cdots, \bm{x}^{(q)}$ to obtain 
	 $F(\bm{x}^{(1)}), \cdots, F(\bm{x}^{(q)})$.
	\STATE augment data $\mathcal{D}_n = \mathcal{D}_{n-1} \cup \{ (\bm{x}^{(1)}, F(\bm{x}^{(1)})), \cdots, $  
	  $(\bm{x}^{(q)}, F(\bm{x}^{(q)})) \}$.   
\ENDWHILE
\end{algorithmic}
\end{algorithm}